\documentclass[letterpaper, 10 pt, conference]{ieeeconf}  % Comment this line out if you need a4paper

\IEEEoverridecommandlockouts                              % This command is only needed if 
\usepackage{cite}
\usepackage{commath}
\usepackage{graphicx}
\usepackage{caption}
\graphicspath{ {Figures/} }

\usepackage{subfig}

\title{\LARGE \bf
Angle-Encoded Swarm Optimization for UAV Formation Path Planning  
}

\author{V.T. Hoang, M.D. Phung, T.H. Dinh, Q.P. Ha % <-this % stops a space
\thanks{The authors are with School of Electrical and Data Engineering, Faculty of Engineering and Information Technology (FEIT), University of Technology Sydney (UTS), 81 Broadway, Ultimo NSW 2007, Australia
        {\tt\small \{VanTruong.Hoang, ManhDuong.Phung, TranHiep.Dinh, Quang.Ha\}@uts.edu.au}}%
}

\begin{document}
\bibliographystyle{IEEEtran}

\maketitle
\thispagestyle{empty}
\pagestyle{empty}

%%%%%%%%%%%%%%%%%%%%%%%%%%%%%%%%%%%%%%%%%%%%%%%%%%%%%%%%%%%%%%%%%%%%%%%%%%%%%%%%
\begin{abstract}
This paper presents a novel and feasible path planning technique for a group of unmanned aerial vehicles (UAVs) conducting surface inspection of infrastructure. The ultimate goal is to minimise the travel distance of UAVs while simultaneously avoid obstacles, and maintain altitude constraints as well as the shape of the UAV formation. A multiple-objective optimisation algorithm, called the Angle-encoded Particle Swarm Optimization ($\theta$-PSO) algorithm, is proposed to accelerate the swarm convergence with angular velocity and position being used for the location of particles. The whole formation is modelled as a virtual rigid body and controlled to maintain a desired geometric shape among the paths created while the centroid of the group follows a pre-determined trajectory. Based on the testbed of 3DR Solo drones equipped with a proprietary Mission Planner, and  the Internet-of-Things (IoT) for multi-directional transmission and reception of data between the UAVs, extensive experiments have been conducted for triangular formation maintenance along a monorail bridge. The results obtained confirm the feasibility and effectiveness of the proposed approach.


\end{abstract}

%%%%%%%%%%%%%%%%%%%%%%%%%%%%%%%%%%%%%%%%%%%%%%%%%%%%%%%%%%%%%%%%%%%%%%%%%%%%%%%%
\section{INTRODUCTION}

Unmanned aerial vehicles (UAVs) have been found in many applications, from farm monitoring to civil infrastructure inspection, logistics to surveillance and rescue, from military to industrial applications with  numerous studies available in the literature, see, e.g. \cite{bri2008, Derafa2012, hayat2016,  Raj2015, mot2016}. However, the increasing demands in applications together with rapid development of technologies, especially computing, sensors, and communications have transcended the use of a single UAV to the formation and coordination of a group of them. Those UAVs in cooperation will be able to accomplish more challenging tasks like drag reduction, telecommunication relay and source and seeking in more efficient ways \cite{gupta2016, Nguyen2004,liao2017}. 

In multiple UAVs formation, path planning and shape preserving are essential for the success of tasks assigned as they provide references for lower level sensing and controlling as well as decide the usability of mission data collected. Over the last decade, there has been a number of studies on this topic with several algorithms introduced. The popular ones include $A^*, D^*$, Rapidly Exploring Random Tree (RRT) and Probabilistic Roadmap Method (PRM) \cite{de2013, yan2013,kothari2013}. The advantage of the $A^*$ and $D^*$ algorithms lie in the ability to judge or evaluate the best point-to-point path so they can provide a flyable trajectory for UAVs \cite{de2013}. The RRT and PRM are based on probabilistic reasoning, and hence, robust but they require the discretization of the operating space which affects the smoothness of the planned trajectory \cite{yan2013,kothari2013}. In \cite{pehlivanoglu2012}, the approach based on  the Voronoi diagram has been developed with a good obstacle avoidance capability. Similarly, path planning based on the visibility binary tree algorithm was introduced in \cite{rashid2013}. In another attempt, the Fast Marching Square (FM2) algorithm \cite{garrido2013, arismendi2015, alvarez2015,  gonzalez2017} has been verified for obtaining excellent results on mobile vehicles, but kinematic constraints have not been considered to validate the feasibility of the trajectories. The Dubins path was introduced as a suitable method to solve this problem in 2D \cite{sujit2008, yu2013} and 3D \cite{chitsaz2007} environments. However, the resulting paths are not optimized. To this end, advances in machine learning have been exploited to provide better results for trajectory generation as in \cite{duan2015, zhang2015}. Those approaches however require large training data as the input. The path planning problem therefore remains  important, especially for coordination of UAV formation in executing a field task \cite{chen2015}. 

In this study, we aim to develop a path planning algorithm for multi-quadcopter formation conducting inspection tasks of built infrastructure. Our approach begins with a 3D model of the inspected surface and its surrounding environmental features extracted from a satellite map. Reference waypoints are then determined. A multi-objective particle swarm optimization algorithm using angle-encoded PSO ($\theta$-PSO) is developed to generate a desired trajectory, for the centroid of the group, that is finally translated into an individual track for each UAV based on its defined position in the formation. Different from path planning for a single UAV \cite{phung2017}, the advantages of the proposed approach rest with not only the simple implementation for the whole group and realistic execution of the planning algorithm, but also the generation of optimized paths for each UAV in the formation. Those paths should be both safely flyable and feasible in order to inspect the structure with the UAV dynamic constraints considered. The paths are also required to pass through pre-defined waypoints. Besides, some new constraints are introduced to improve the collision avoidance capability and task efficiency. Extensive simulation, comparison and experiments have been conducted for evaluation. The results illustrate the validity and effectiveness of the proposed path planning algorithm. 

The paper is organized as follows. Section \ref{model} briefly describes the model of the quadcopter formation. Section \ref{F_planning} presents the design of the path planning algorithm using $\theta$-PSO. Section \ref{I_planning} describes the implementation of trajectory planning for UAVs. Simulation and experimental results are introduced in Section \ref{results}. The paper ends with a conclusion and recommendation for future work.

%%%%%%%%%%%%%%%%%%%%%%%%%%%%%%%%%%%%%%%%%%%%%%%%%%%%%%%%%%%%%%%%%%%%%%%%%%%%%%%%
\section{TRIANGULAR FORMATION MODEL} \label{model}
\begin{figure}[!htbp]
\centering
	\vspace{-3mm}
	\includegraphics[width=7cm]{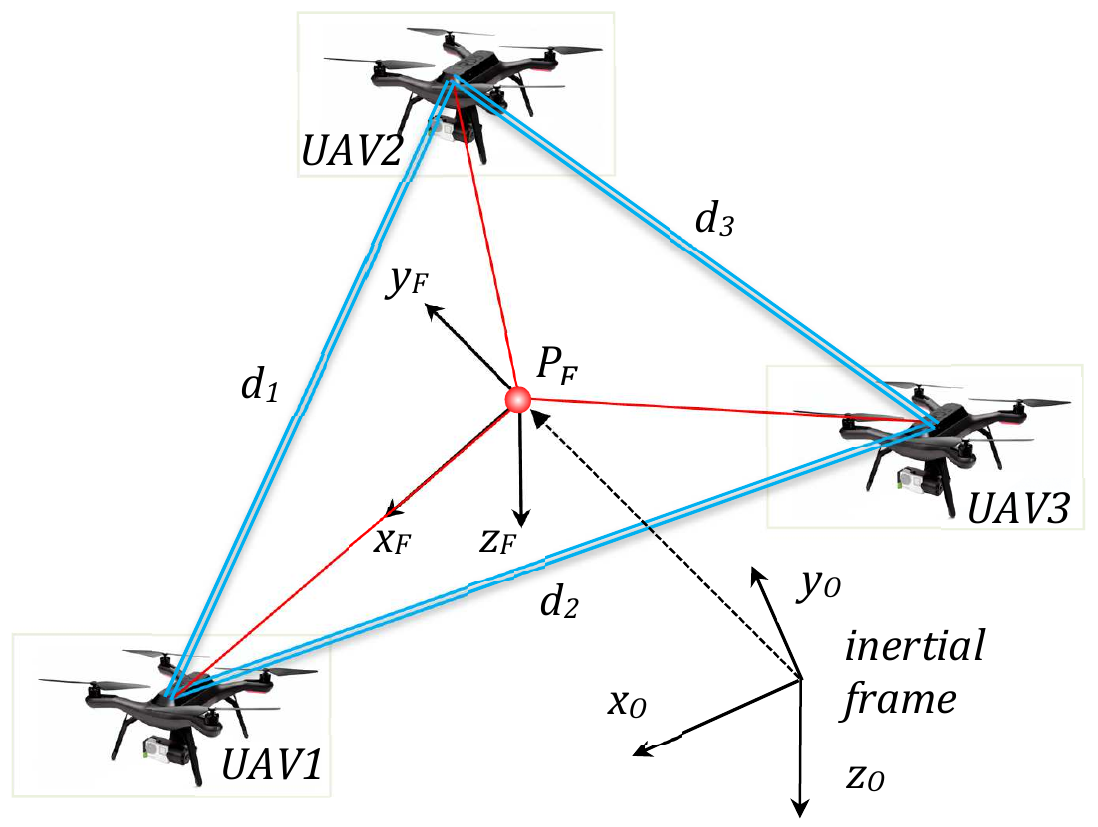}
	\caption{Inertial and formation frames in UAV formation}
	\label{form_fig}
	\vspace{-3mm}
\end{figure}

%\subsection{Formation configuration structure}
Figure \ref{form_fig} shows the inertial and formation frames that represent a triangular UAV formation. All measurements are referred to the inertial frame $O$ with axes $x_O, y_O$ and $z_O$.  Positions of UAV$_n$, $n=1,2,3$, in the inertial frame are denoted as $P_n = \{x_n, y_n,z_n\}$. The formation frame, $\{x_F, y_F, z_F\}$, is defined such that the origin $P_F$ is chosen to be coincident with the centroid of the triangle; the axis $x_F$ is the direction from the centroid of the triangle to the UAV$_1$ position; the axis $z_F$ is perpendicular to the plane containing three UAVs pointing downward; and the axis $y_F$ is perpendicular to the plane formed by the $x_F$ and $z_F$ axes. This moving frame allows to determine its relative orientation with respect to the fixed inertial frame. 
The group position is represented by the centroid of the triangle and expressed as:
\begin{equation}
P_F = \dfrac{1}{3} \sum\limits_{i=1}^3 P_n.
\end{equation}
Let denote the distances between UAV$_1$, UAV$_2$ and UAV$_3$ to the centroid of the triangle as $d_{1,F}$, $d_{2,F}$, and $d_{3,F}$, respectively. The rotation matrix which represents the relation between the formation and inertial frames is determined as:
\begin{align}\label{Eq_rotation_matrix2}
R_{OF} =  \left [\begin{array}{ccc} 
c_{\psi}c_{\theta} & c_{\psi}s_{\theta} s_{\phi}-s_{\psi}c_{\phi} & c_{\psi}s_{\theta} c_{\phi}+s_{\psi}s_{\phi}\\
s_{\psi}c_{\theta} & s_{\psi}s_{\theta} s_{\phi}+c_{\psi}c_{\phi} & s_{\psi}s_{\theta} c_{\phi}-c_{\psi}s_{\phi}\\
-s_{\theta} & c_{\theta} s_{\phi} & c_{\theta} c_{\phi}
\end{array} \right],
\end{align}
\noindent where $s_x = sin(x)$, $c_x = cos(x)$, and $\phi$, $\theta$ and $\psi$ are Euler angles of the shape. 
%\subsection{Formation motion model}
%%%%%%%%%%%%%%%%%%%%%%%%%%%%%%%%%%%%%%%%%%%%%%%%%%%%%%%%%%%%%%%%%%%%%%%%%%%%%%%%
\section{FORMATION PATH PLANNING} \label{F_planning}
When producing a path for the desired motion of multiple UAVs in a group, a number of constraints are required to be fulfilled for maintenance of the formation, maneuverability of every single UAV, operating space, and obstacle avoidance. In this work, all of the constraints will be incorporated into a multi-objective function. The path planning problem can be then simplified to the creation of a feasible path for the centroid of the UAV formation. Since our goal is to construct optimal paths for all UAVs in the group, it is essential to speed up the convergence of the optimization process for the whole formation. Therefore, we propose to use the angle-coded PSO ($\theta$-PSO) described as follows.
\subsection{Angle-encoded PSO or $\theta$-PSO}
%\subsubsection{Principle of the PSO Algorithm}
The PSO is a population-based stochastic optimization algorithm inspired by social behavior of bird flocking \cite{shi1998,fu2012}. In PSO, a set of particles is generated, each seeks for the optimum solution by moving in a way that compromises between its own experience and the social experience. Initially, each particle is assigned a random position, $x_i$, and velocity, $v_i$. The particle motion is then updated by the following equations:
\begin{align}\label{EqPSO}
v_{ij}^{k+1} &= w v_{ij}^k + c_1 r_{1i}^k (p_{ij}^k - x_{ij}^k) + c_2 r_{1i}^k (p_{gj}^k - x_{ij}^k) \\
x_{ij}^{k+1} &= x_{ij}^k + v_{ij}^{k+1}, (i=1,2,...,N; j=1,2,...,S),
\end{align}
where $N$ is the swarm size, $S$ is dimension of the searching space, $w$ is the inertial weight, $r_1$ and $r_2$ are two pseudorandom scalars, $c_1$ and $c_2$ are the gain coefficients, $p_{ij}$ and $p_{gj}$ are the local-best and global-best positions of the particle $i$, and subscript $k$ is the iteration index. The values of $p_{ij}$ and $p_{gj}$ are evaluated based on a cost function to be defined in the next section.

For traditional path planning, the position of particles often represents the location of UAVs. This representation can give good results, but it also slows down the swarm convergence if the momentums of particles are not well adjusted \cite{kwok2007}. To overcome this problem, we propose to use the angle-encoded PSO or $\theta$-PSO, motivated by \cite{fu2012}, in which the location of particles is encoded by the angle of UAVs. The $\theta$-PSO is described as:
\begin{align}\label{EqThePSO}
\begin{cases} \Delta \theta_{ij}^{k+1} = w \Delta \theta_{ij}^k + c_1 r_{1i}^k (\lambda_{ij}^k - \theta_{ij}^k) + c_2 r_{1i}^k (\lambda_{gj}^k - \theta_{ij}^k) \\
\theta_{ij}^{k+1} = \theta_{ij}^k + \Delta \theta_{ij}^{k+1}, (i=1,2,...,N; j=1,2,...,S)\\
x_{ij}^k = \dfrac{1}{2}\left[(x_{max} -x_{min}) sin \left( \theta_{ij}^k \right) + x_{max} + x_{min} \right], \end{cases}
\end{align}
where $\theta_{ij} \in [-\pi/2, \pi/2]$ and $\Delta \theta_{ij} \in [-\pi/2, \pi/2]$ are respectively the phase angle and phase angle increment of the $i$th particle in dimension $j$; $\Lambda_g = [\lambda_{g1}, \lambda_{g2},..., \lambda_{gS}]$ and $\Lambda_i = [\lambda_{i1}, \lambda_{i2},..., \lambda_{iS}]$  are respectively the global and personal best positions; and $x_{max}$ and $x_{min}$ are the upper and lower restrictions of the search space.

For path planning of the centroid, each particle is associated to a specific path instance $T_{Fi}$ and the phase angle-encoded population can be presented as:
\begin{equation}
\Theta = [\Theta_1,\Theta_2, ...,\Theta_N]^T
\end{equation}
Suppose each path $T_{Fi}$ consists of $v + 2$ waypoints, including the start and target ones. As those start and target waypoints are predetermined, they can be excluded from the particle. Thus each particle has the dimension of $3v$ and can be represented by the following fixed-length phase angle-encoded vector: 
\begin{equation}
\Theta_i = [\theta_{i1}, ..., \theta_{iv}, \theta_{i,v+1}, ..., \theta_{i,2v}, \theta_{i,2v+1}, ..., \theta_{i,3v}]. 
\end{equation}
Using the mapping $f:[-\pi/2, \pi/2] \rightarrow [x_{min}, x_{max}]$, we obtain the position of a particle as:
\begin{equation}
X_i=f(\Theta_i) = [x_{i1}, ..., x_{iv}, x_{i,v+1}, ..., x_{i,2v}, x_{i,2v+1}, ..., x_{i,3v}],
\end{equation}
where $x_{ij} = f(\theta_{ij})$ is the $j$th dimension of the $i$th particle's position $(i = 1,..., N$; $j = 1,..., 3v)$, $x_{i1},..., x_{iv}, x_{i,v+1},..., x_{i,2v}$ and $x_{i,2v+1},..., x_{i,3v}$ represent the $x$, $y$ and $z$ coordinates of the $v$th waypoint of path $T_{Fi}$, respectively.
\subsection{Cost Function}
The selection of a proper cost function for the PSO is essential to achieve the globally optimal solution for the searching process. The cost function for any trajectory often forms by two major evaluations, the length and violation cost of the path. The former helps to minimize the total travelling distance of the path whereas the latter is to avoid collisions of UAVs with each other and with obstacles. In a 3D site, other constraints should also be required, e.g., restrictions in flying altitude, heading angle and path curve. In our system, we use the quadcopters that have capabilities to carry out sharp and abrupt changes in angles and curves \cite{Hoang2017AT}. Thus, the constraints in angle and curve can be relaxed. The multi-objective constraint is now incorporated into the cost function in the following form:
\begin{equation} \label{cost_function}
J_F(T_{Fi}) = \sum \limits_{n=1}^3 \beta_n J_n(T_{Fi}),
\end{equation}
where $T_{Fi}$ is the formation path; $\beta_n$ is the weighting factor indicating the corresponding threat intensity; and $J_n(T_{Fi})$, $n=1,2,3$, are the costs associated with the path length, collision violation and flying altitude, respectively. To determine $J_n(T_{Fi})$, we split the formation path $T_{Fi}$ into $m$ segments. Each segment is represented by coordinates of its ending nodes $P_{Fi,l} = \{x_{Fi,l},y_{Fi,l},z_{Fi,l}\}, l= 0 .. m$. By denoting the length of the segment connecting nodes $P_{Fi,l}$ and $P_{Fi,l+1}$ as $\norm{\overrightarrow{P_{i,l}P_{i,l+1}}}$, the cost $J_1$ corresponding the path length is then calculated for all segments:
\begin{align}
J_1(T_{Fi}) &= \sum \limits_{l=0}^m \norm{\overrightarrow{P_{i,l}P_{i,l+1}}}\\
&= \sum\limits_{l=0}^m \sqrt{dx_{Fi,l}^2 + dy_{Fi,l}^2 + dz_{Fi,l}^2} .
\end{align}

\begin{figure}[!htbp]
	\centering
	\vspace{-3mm}
	\includegraphics[width=6cm]{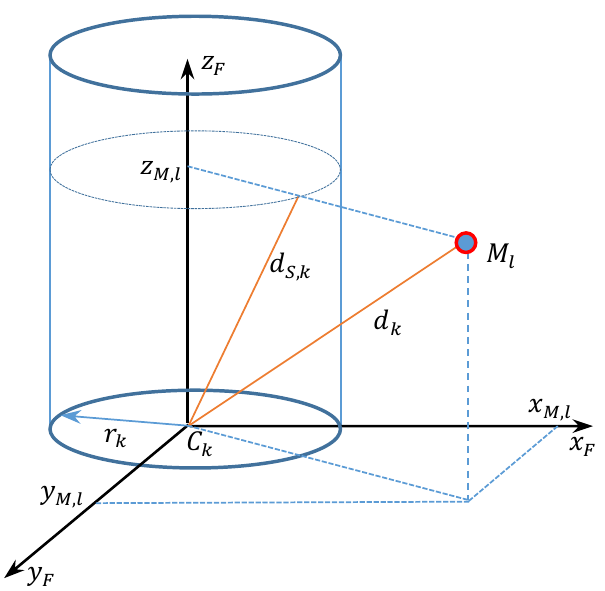}
	\caption{Obstacle representation and safe distance calculation}
	\label{obs_safe}
	\vspace{-3mm}
\end{figure}

Let $K$ be the set of all obstacles for a given UAV within its operation space. Assume that each obstacle is prescribed in a cylinder with the center's coordinate $C_k$ and radius $r_k$, as shown in Fig. \ref{obs_safe}. The surfaces of cylinders then can be used to form constraints for obstacle avoidance. Specifically, the safe distance $d_{s,k}$ from the obstacle $k$ is calculated from the cylinder center to its surface at the altitude $z_{M,l}$ as:
\begin{equation}
d_{s,k} =  \begin{cases} \sqrt{r_k^2 + (z_{M,l} - z_k)^2}  &\text{ if }  z_{M,l} \le z_{max, k}\\
\sqrt{r_k^2 + (z_{max,k} - z_k)^2}   &\text{ if }  z_{M,l} > z_{max, k},
\end{cases} 
\end{equation}
where $z_{max,k}$ is the height of obstacle $k$. 
To compute the violation cost between each generated path and obstacle centres, we first assumed that the formation is rigid and can be fit within a sphere with the radius
\begin{equation}
r_{s,k} = r_Q + r_F + d_{s,k}, 
\end{equation}
where $r_Q$ is the radius of quadcopters including propellers, $r_F$ is the radius of the formation,
\begin{equation}
r_F = \text{max}(D_{i,F}),
\end{equation}
in which $D_{i,F}$ is the distance from quadcopter $i$ to the formation centroid. The violation cost can be now derived as follows: 

For the $k$th obstacle, compute the distance from its center $C_k$ to the segment $\overrightarrow{P_{i,l}P_{i,l+1}}$:
\begin{equation}
d_{l,k} =  \sqrt{(x_{M,l} - x_k)^2 + (y_{M,l} - y_k)^2 + (z_{M,l} - z_{k})^2},
\end{equation}
where $M_l = \{ x_{M,l},y_{M,l},z_{M,l}\}$ is the midpoint of the segment as shown in Fig. \ref{obs_safe}. At a given altitude $z_{M,l}$, $d_{l,k}$ is then compared with the safe distance to the obstacle. The comparison results in the following violation function:
\begin{equation}
V_{l,k}(T_{Fi}) = \sum \limits_{k=1}^K \text{max} (1-\dfrac{d_{l,k}}{r_{s,k}},0).
\label{violation}
\end{equation}
This function ensures that the distance $d_{l,k}$ must be larger than the safe distance for obstacle avoidance. The violation cost is then computed for  all obstacles as:
\begin{equation}
V_{l}(T_{Fi}) = \frac{1}{K} \sum \limits_{k=1}^K V_{l,k}(T_{Fi}).
\end{equation}
For all $m$ segments, the final violation cost on average (with respect to the centroid) is represented as:
\begin{equation}
J_2(T_{Fi}) =  \frac{1}{m}\sum \limits_{l=0}^m V_l(T_{Fi}),
\end{equation}
In terms of flying altitude, UAVs are often required to follow the terrain at a certain height to avoid crashing. Thus, the altitude of each UAV must be within a predefined interval between two given extrema, the minimum and maximum safe clearances $z_{\text{min}}$ and $z_{\text{max}}$. Thus, the corresponding cost component can be expressed as
\begin{align}
\begin{cases}
&J_3(T_{Fi}) =  \sum \limits_{l=0}^{m} dz_{Fi,l}\\
&dz_{Fi,l} = \begin{cases}
			z_{Fi,l} - z_{\text{max}},  &\text{if } z_{Fi,l} > z_{\text{max}}\\
			0,  &\text{if } z_{\text{min}} \leq z_{Fi,l} \leq z_{\text{max}}\\
			z_{\text{min}} - z_{Fi,l},  &\text{if } 0< z_{Fi,l} < z_{\text{min}}\\
			\infty,  &\text{if } z_{Fi,l} \leq 0.
			\end{cases}
\end{cases}
\end{align}
%The inequality becomes an equality only in the case that UAVs are on the ground. 
This last condition in $dz_{Fi,l}$ is critical for safe operations as negative values of the altitude could be generated causing UAVs to crash to the ground.

\subsection{Path planning implementation}
\begin{figure}[!htbp]
	\centering
	\vspace{-3mm}
	\includegraphics[width=8cm]{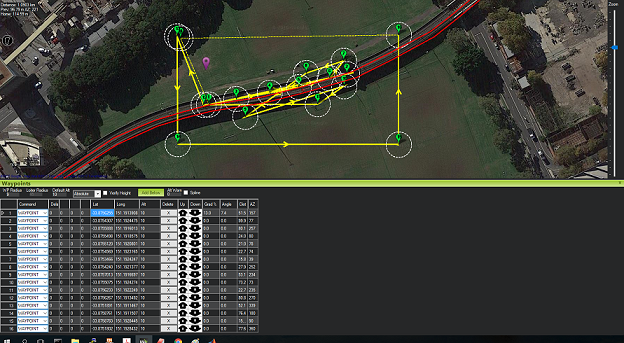}
	\caption{Mission Planner incorporating Google Satellite Map to create initial information and an inspection plan}
	\label{WW_Bridge_plan}
	\vspace{-3mm}
\end{figure}

The implementation starts with choosing the operation space of UAVs and the infrastructure to be inspected. This can be done by using a navigation map with satellite images. For example, here a monorail bridge as a testbed subject to inspection can be loaded on the Mission Planner, as shown in Figure \ref{WW_Bridge_plan}. The obstacles are also identified based on this map. Furthermore, range sensors such as lidars can be used to form a 3D map of the environment as in our previous work \cite{phung2017}. Based on those inputs, the cost function together with constraints can be defined as described in the previous section. The $\theta$-PSO algorithm will then be run to obtain the desired path. 
The steps can be summarized as follows:
\begin{itemize}
\item[(1)] Compute all parameters for UAVs and formation model, based on the desired task.
\item[(2)] Select appropriate parameters for PSO such as the population size, the number of iterations, gain coefficients and so on. 
\item[(3)] Identify environmental information of the flying field and initialise parameters of obstacles.
\item[(4)] Initialize randomly the path of each particle from the start point to the target point.
\item[(5)] Evaluate each path based on the cost function (\ref{cost_function}).
\item[(6)] Compute each particle's personal best and the global best positions by running PSO repeatedly.
\item[(7)] The desired path is chosen as the maximum number of iterations is reached.
\end{itemize}

%%%%%%%%%%%%%%%%%%%%%%%%%%%%%%%%%%%%%%%%%%%%%%%%%%%%%%%%%%%%%%%%%%%%%%%%%%%%%%%%
\subsection{Path generation for individual UAV} \label{I_planning}
Given the optimal path, $T^*_F$, generated by the $\theta$-PSO for the formation centroid, it is necessary to produce a specific path for each UAV so that the shape of the formation during the flight can be maintained. Those paths can be computed based on the PSO's generated path and the desired relative distances among the UAVs. Let $P_{n,d} = [x_{n,d} , y_{n,d} , z_{n,d}]^T$ be the reference position for each UAV during the flight and $P_n = [x_n, y_n, z_n]^T$ be the actual position, $P_n$ can be obtained from GPS data of the $n$th UAV. We then define the relative position errors of the $n$th UAV during the flight in the inertial frame as:
\begin{align} \label{eq_error}
\left[ \begin{array}{c}
e_{nx} \\e_{ny} \\e_{nz}
\end{array} \right]  = \left[ \begin{array}{c}x_{n,d} -x_n\\ y_{n,d} - y_n \\ z_{n,d} - z_n\end{array} \right] 
\end{align}
Using the rotation matrix (\ref{Eq_rotation_matrix2}), with $R_{FO}(t) = R^{-1}_{OF}(t)$, the errors in (\ref{eq_error}) can be converted into the errors in the formation frame as:
\begin{align} \label{eq_error2}
\left[ \begin{array}{c}
e_{nxF} \\e_{nyF} \\e_{nzF}
\end{array} \right]  = R_{FO}(t)\left[ \begin{array}{c} e_{nx} \\e_{ny} \\e_{nz} \end{array} \right] 
\end{align} \label{indi_path}
The customized path for each UAV can then be represented in terms of trajectory control command as: 
\begin{equation}
T_n = T^{*}_{F} + \Delta T_n,
\end{equation}
where $T^{*}_{F} $ is the trajectory of the centroid, computed by $\theta$-PSO, and $\Delta T_n$ is the amount added to direct the UAV away from the centroid. $\Delta T_n$ is calculated from the desired relative distances among the UAVs and the relative position errors in (\ref{eq_error2}). The output $T_n$ will be fed to the internal controller of UAVs for trajectory tracking.

%%%%%%%%%%%%%%%%%%%%%%%%%%%%%%%%%%%%%%%%%%%%%%%%%%%%%%%%%%%%%%%%%%%%%%%%%%%%%%%%
\section{EXPERIMENTS} \label{results}
A series of experiments have been conducted to evaluate the feasibility and efficiency of the proposed algorithm.
\subsection{Experimental setup}
The task assigned in experiments is to inspect simultaneously different surfaces of a bridge using three UAVs. As mentioned, the Mission Planner, a proprietary ground control station software, incorporating the Google Satellite Map (GST) is used to collect initial information about the structure and its surrounding environment. Here, the operation space is chosen with dimensions $\{141 m \times 101 m \times 40 m\}$, equivalent to the GST coordinates $\{ -33.87601, 151.191182, 0 \}$ and $\{-33.875086, 151.192676, 40\}$, as illustrated in Fig. \ref{WW_Bridge_plan}.  The starting and target points are set as $P_{i,0} = \{40.0,8.0,30\}$ and $P_{i,l+1} =\{64,108,34\}$, respectively. Therein, ten obstacles are identified, each with a different radius. 
\begin{figure}[!htbp]
	\vspace{-3mm}
	\centering
	\includegraphics[width=5cm]{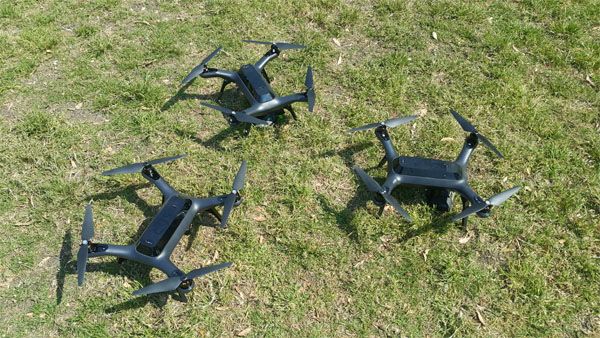}
	\caption{3DR Solo drones used in experiments}
	\label{drone_fig}
	\vspace{-3mm}
\end{figure}
The formation platform chosen in this work is three identical 3DR Solo drones as shown in Fig. \ref{drone_fig}, whereby the controllers at the low level for each UAV have been reported in \cite{Hoang2017AT}. The communications among them were conducted by adding an additional Internet-of-Things (IoT) board to each drone and a base Wi-Fi router. The IoT boards together with the ground Wi-Fi station form a network that can connect to the internet to transmit the data to other processing station. Also through this network, the drones can exchange their position, velocity and status data during flight. By employing that information, the onboard computer calculates the inverse kinematics (the formation variables based on the positions of the robots), compares it with their neighbours and the formation centroid to obtain the position errors. Those errors are then eliminated by the tracking control action generated.
\begin{figure}[!htbp]
	%\vspace{-3mm}
	\centering
	\includegraphics[width=8cm]{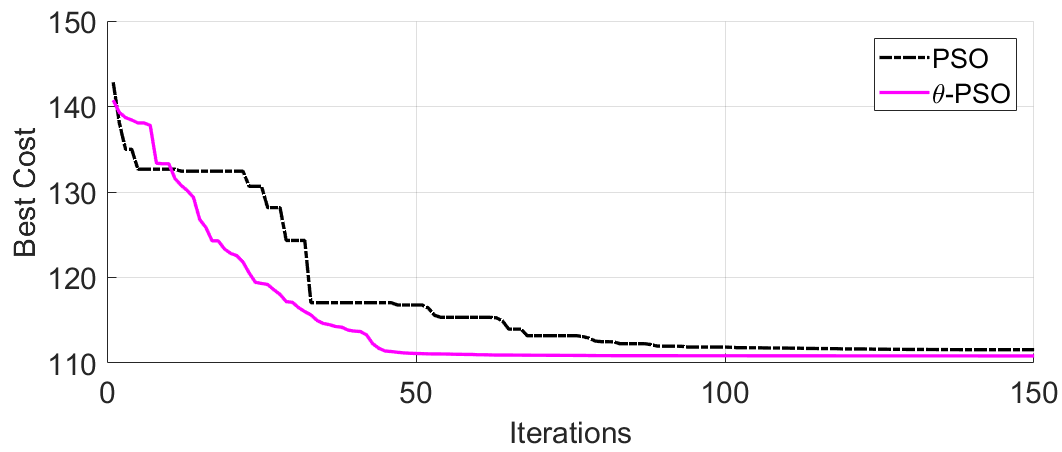}
	\caption{Convergence comparison between conventional PSO and $\theta$-PSO}
	\label{Comparison}
	\vspace{-3mm}
\end{figure}
In our path planning algorithm, the number of particles, waypoints, and iterations are respectively selected as 100, 10, and 300. Parameters of the three quadcopters with respect to the centroid of the formation are $\Delta T_1 = [0, 0, 2]$ m, $\Delta T_2 = [3, 0, -1]$ m and $\Delta T_3 = [-3, 0, -1]$ m. The minimum and maximum clearances between UAVs and the terrain are set to $z_{\text{max}} = 32$ m and $z_{\text{min}} = 28$ m, respectively. 
\subsection{Results}
The path planning results are presented in this subsection, whereby it is expected that the designed method can generate collision-free paths for the three UAVs with sufficiently fast convergence using the proposed algorithm. For this, let us first compare the performance of the proposed $\theta$-PSO with a conventional PSO algorithm. Figure \ref{Comparison} shows the cost values over iterations. It can be seen that although both algorithms are convergent, the $\theta$-PSO introduces a faster and more stable conversion. The results are confirmed as recorded in Table \ref{Table2} which shows the average cost value and convergence iterations.
\begin{center}
	\captionof{table}{PSO and $\theta$-PSO performance comparison} \label{Table2} 
	\begin{tabular}{ |c|c|c|c|}
		\hline
		Algorithm	& Min cost 		& Max cost 	& Iterations \\ \hline
		PSO 		& 112.43	 		& 143.0	& 102	\\
		$\theta$-PSO & 111.02			& 142.84	& 68\\
		\hline 
	\end{tabular}
\end{center}
\begin{figure}[!t]
	\vspace{-3mm}
	\centering
	\subfloat[Triangular UAV formation]{
	\includegraphics[width=6cm]{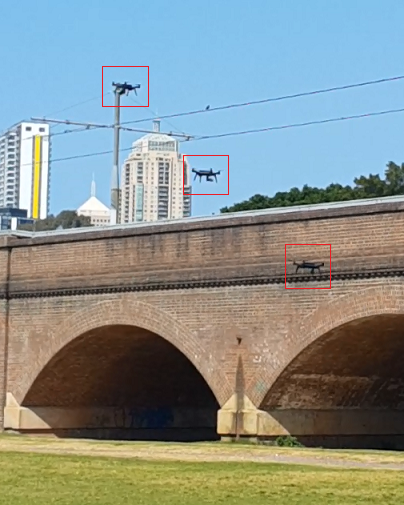}
	\label{Solo_form_fig}
	} 
\hfil
	\centering
	\subfloat[Planned path (yellow) and flown path (violet)]{
	\includegraphics[width=7cm]{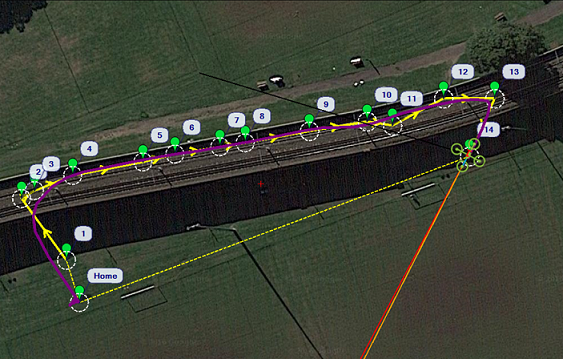}
	}
		\caption{Bridge inspection with UAV formation}
	\vspace{-3mm}
		\label{fig6}
\end{figure}
Given the generated path, we have conducted field-test experiments in which the triangular formation automatically navigated along the inspected surface, as depicted in Fig. \ref{fig6}. The 3D trajectories generated from that of the formation centroid path are shown in Fig. \ref{path_3Solo}, where it can be seen that the three drones can take off, reach to their individual altitude set-point, descend and finally arrive their target position at almost the same interval, while maintaining the desired triangular shape. This result can be further verified via the altitude time responses of the three UAVs as recorded in Fig. \ref{Alt_Solo_Formation}. It is clear that the quadcopters are capable of avoiding obstacles and preserving the desired formation configuration during the inspection task. For further evaluation, Fig. \ref{Path_error} shows the error between the planned and flown paths computed by selecting the closest coordinates of the flight trajectories to the reference points. Those small errors, mainly caused by the positioning system of UAVs from the GPS signal received, imply feasibility and reasonable smoothness of the generated path for the deployment of the drone formation.
\begin{figure}[!htbp]
	\vspace{-3mm}
	\centering
	\includegraphics[width=8cm]{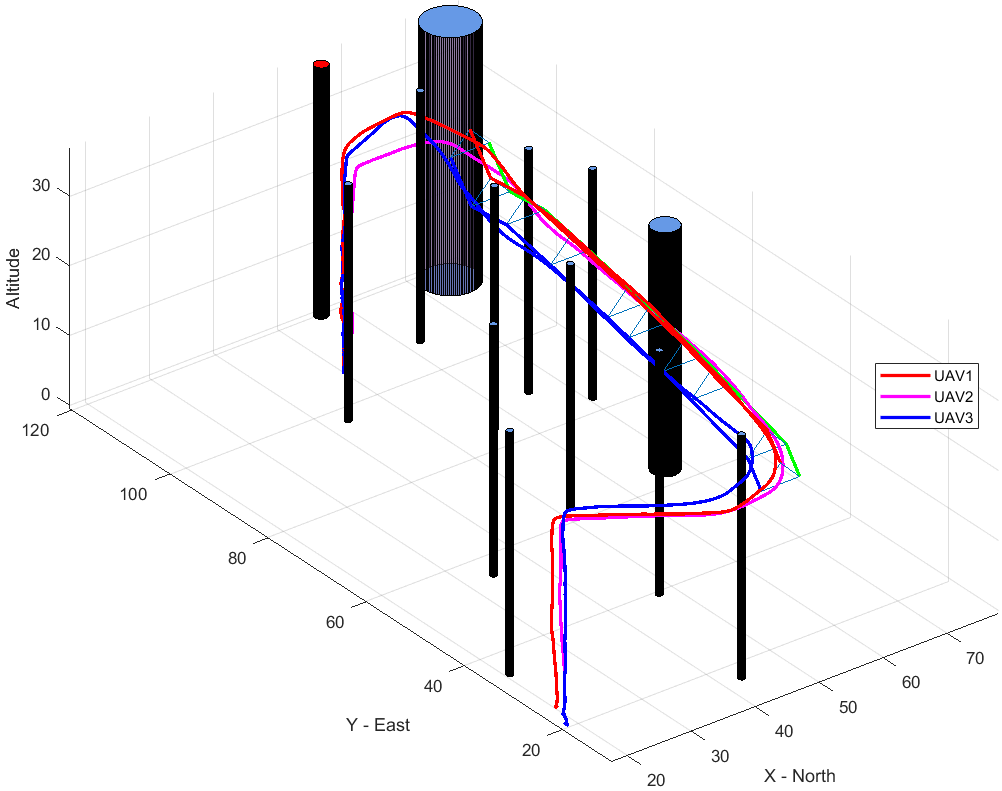}
	\caption{Trajectories of three drones tracking the planned paths}
	\label{path_3Solo}
	\vspace{-3mm}
\end{figure}

\begin{figure}[!htbp]
%	\vspace{-3mm}
	\centering
	\includegraphics[width=8cm]{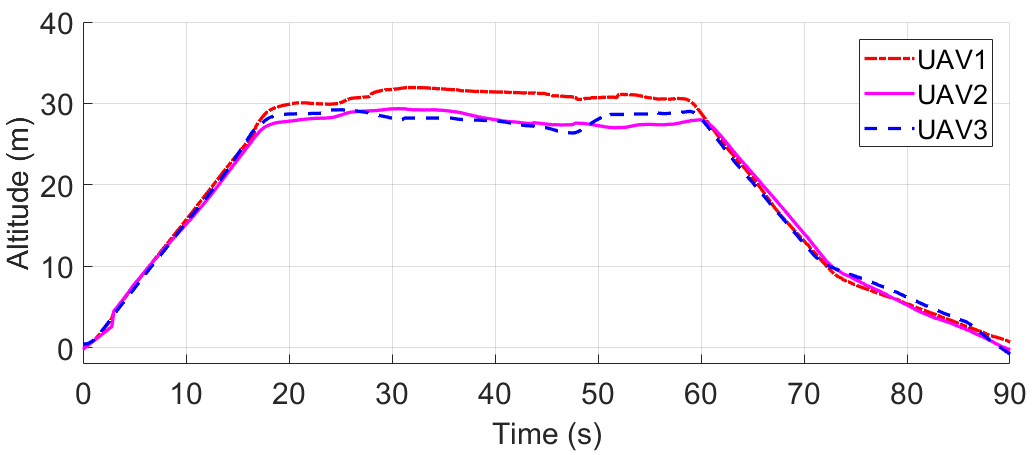}
	\caption{Altitudes of the three drones in the formation test}
	\label{Alt_Solo_Formation}
	\vspace{-3mm}
\end{figure}

\begin{figure}[!htbp]
%	\vspace{-3mm}
	\centering
	\includegraphics[width=8cm]{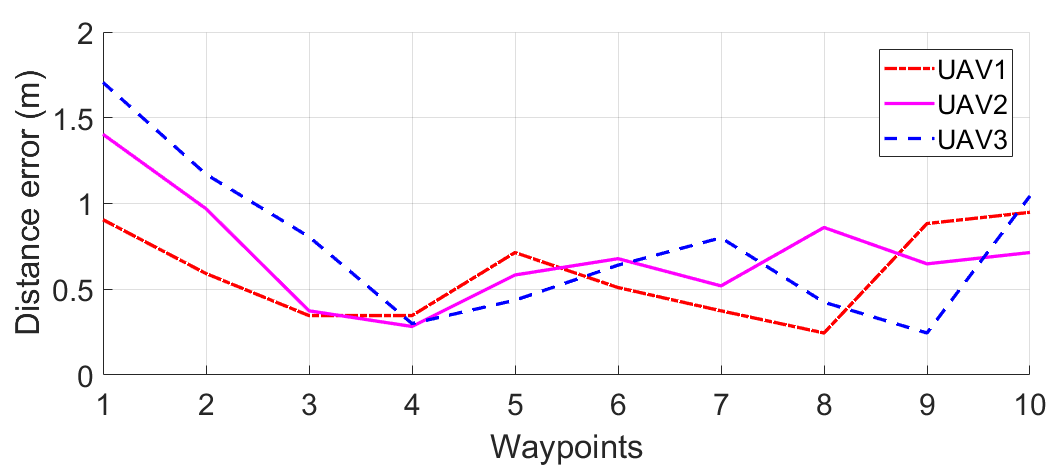}
	\caption{Errors between the planned and flown paths}
	\label{Path_error}
	\vspace{-3mm}
\end{figure}
%%%%%%%%%%%%%%%%%%%%%%%%%%%%%%%%%%%%%%%%%%%%%%%%%%%%%%%%%%%%%%%%%%%%%%%%%%%%%%%%
\section{CONCLUSION}
This paper has presented a novel approach for the path planning problem of multiple UAVs navigating in a desired shape for infrastructure inspection tasks. Here the angle-coded PSO is proposed to find feasible and obstacle-free paths for the whole formation by minimizing a cost function that incorporates multiple constraints for shortest paths and safe operation of the drones. From the centroid, customized paths are generated for individual UAV to maintain the formation using a proprietary software while inter-UAV communication is achieved via the IoT boards. Implementation on a triangular formation is reported along with field tests on a monorail bridge. The results confirmed the validity and feasibility of the proposes approach for UAV formation inspection of built infrastructure.

%\addtolength{\textheight}{-14.5cm}   % This command serves to balance the column lengths

%%%%%%%%%%%%%%%%%%%%%%%%%%%%%%%%%%%%%%%%%%%%%%%%%%%%%%%%%%%%%%%%%%%%%%%%%%%%%%%%
\bibliography{IEEEabrv,bibi}

\end{document}